\documentclass{article}

\usepackage{microtype}
\usepackage{graphicx}
\usepackage{subfigure}
\usepackage{booktabs} 

\usepackage{hyperref}

\usepackage[accepted]{icml2023}

\usepackage{amsmath}
\usepackage{amssymb}
\usepackage{mathtools}
\usepackage{amsthm}

\usepackage[capitalize,noabbrev]{cleveref}

\theoremstyle{plain}

\theoremstyle{definition}

\theoremstyle{remark}

\usepackage[textsize=tiny]{todonotes}

\icmltitlerunning{Work in progress.}

\begin{document}

\twocolumn[
\icmltitle{OneCAD: One Classifier for All image Datasets \\
using multimodal learning
}



\icmlsetsymbol{equal}{*}

\begin{icmlauthorlist}
\icmlauthor{Shakti N. Wadekar}{equal,ece}
\icmlauthor{Eugenio Culurciello}{bio}
\end{icmlauthorlist}

\icmlaffiliation{ece}{Department of Electrical and Computer Engineering, Purdue University, West Lafayette, IN, USA}
\icmlaffiliation{bio}{Department of Biomedical Engineering, Purdue University, West Lafayette, IN, USA}

\icmlcorrespondingauthor{Shakti N. Wadekar}{swadekar@purdue.edu}

\icmlkeywords{Machine Learning, ICML}

\vskip 0.3in
]



\printAffiliationsAndNotice{\icmlEqualContribution} 

\begin{abstract}

Vision-Transformers (ViTs) and Convolutional neural networks (CNNs) are widely used Deep Neural Networks (DNNs) for classification task.
These model architectures are dependent on the number of classes in the dataset it was trained on.
Any change in number of classes leads to change (partial or full) in the model's architecture.
This work addresses the question: \textit{Is it possible to create a number-of-class-agnostic model architecture?}.
This allows model's architecture to be independent of the dataset it is trained on.
This work highlights the issues with the current architectures (ViTs and CNNs). 
Also, proposes a training and inference framework \textbf{OneCAD} (\textbf{One} \textbf{C}lassifier for \textbf{A}ll image \textbf{D}atasets) to achieve close-to number-of-class-agnostic transformer model.
To best of our knowledge this is the first work to use Mask-Image-Modeling (MIM) with multi-modal learning for classification task to create a DNN model architecture agnostic to the number of classes.
\textit{Preliminary results} are shown on natural and medical image datasets.
Datasets: MNIST, CIFAR10, CIFAR100 and COVIDx.
Code will soon be publicly available on github.




\end{abstract}

\begin{figure*}[]
\vskip 0.1in
\begin{center}
\centerline{\includegraphics[scale=0.7]{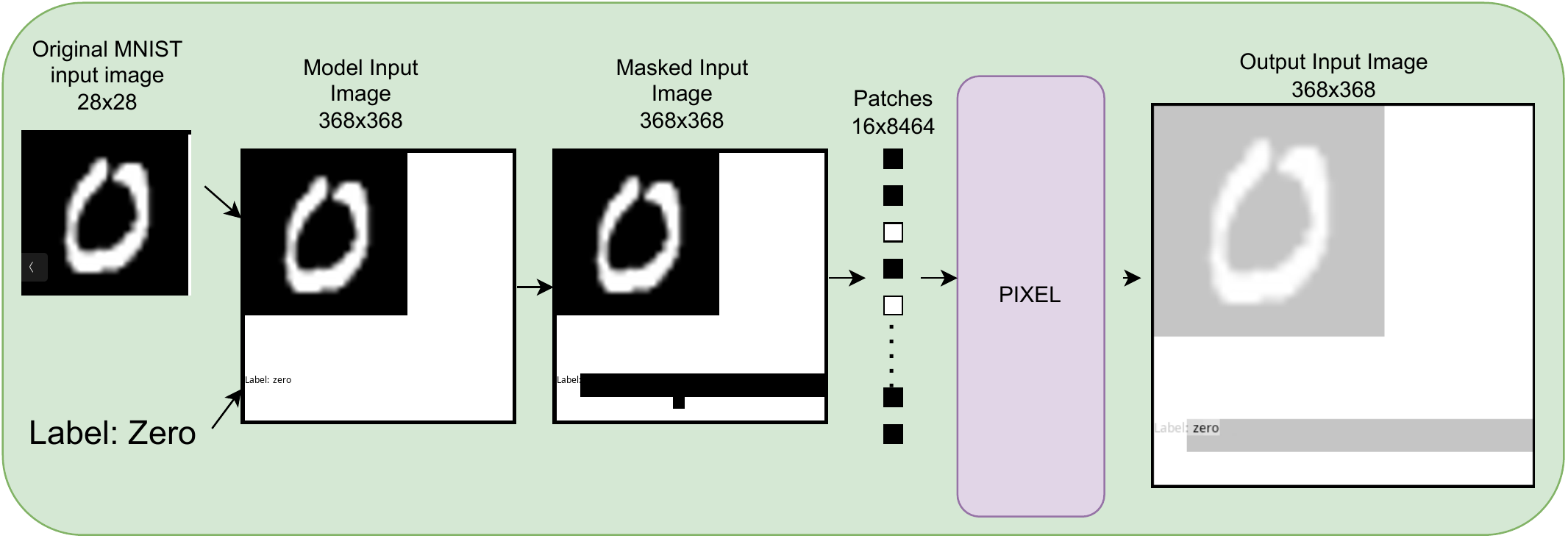}}
\caption{The figure shows our proposed OneCAD's training loop. The input image and its text label is copied on 368x368 image. The label pixels are masked. The PIXEL model takes this masked input image and tries predicts text-pixels matching to the label.}
\label{fig:modeltrain}
\end{center}
\vskip -0.1in
\end{figure*}

\section{Introduction}
\label{submission}
Classification is one of the most widely performed task across different domains.
After the advent of AlexNet \cite{krizhevsky2017imagenet}, a Deep Neural Network (DNN), neural networks have dominated the classification task and various other tasks specially in computer vision domain.
More deeper networks like, VGG \cite{simonyan2014very}, ResNet \cite{he2016deep}, InceptionNet \cite{szegedy2016rethinking}, DenseNet \cite{huang2017densely} and EfficientNet \cite{tan2019efficientnet} have further achieved higher accuracies in this task.
Transformer model, which is extensively used in NLP (Natural Language Processing) domain, is also being applied to vision domain.
ViT \cite{dosovitskiy2020image} was the first transformer based model applied to vision tasks including classification.
Many modified versions of transformer model like, CoAtNet \cite{dai2021coatnet}, CoCa \cite{yu2022coca}, SwinV1 \cite{liu2021swin}, SwinV2 \cite{liu2022swin}, BEiT \cite{bao2021beit}, and MViT \cite{li2022mvitv2} have achieved state of the art results on classification task.

Deep Neural Network models when used for classification task, their architectures depend on the number of classes in the dataset it is trained on.
If the number of classes change, then the model's architecture changes (partially or fully).
Two step procedure is widely used for adapting the model to new class.
First, model's architecture is changed.
Mostly, the last linear layer is changed to have the output equal to new number of classes. 
Second, partial or full retraining of the model is done.

For a true general classification model, it should be able to classify large number of classes if not infinite. 
This is a typical case for online vendors with millions of stock items.
As the number of classes grow, the number of parameters required in this last layer also increases.
This is due to one-hot-encoding of the classes.
Let $f$ be number of output features from the model and $N$ be the number of classes.
Total number of parameters required in the last linear layer (fully connected layer) will be $\textit{f} \times \textit{N}$.
If the model has 4096 output features ($f$) and 10 million classes ($N$), then number of parameters contributed by last linear layer alone would be $\sim$40 billion parameters. 
So using one-hot-encoding (one output of linear layer to represent one class) becomes computationally expensive for large classification task.

This work proposes a solution using multi-modal learning as shown in Figure \ref{fig:modeltrain}.
Masked-Auto-Encoder (MAE) Vision-Transformer model called PIXEL \cite{rust2022language} is used to predict the masked image patches with class name (text) in it.
Representing and predicting the class output by visual-text allows model to predict large number of classes with limited number of pixels/patches.
For example, if one output image patch can represent one english-letter, then it can have 26 different possibilities. If total 10 patches in sequence are allowed to represent the class name, then $\sim$$10^{14}$ i.e, greater than 100 trillion class names/words can be represented. 
If 20 patches are allowed, then $10^{28}$ (ten thousand quadrillion) class names/words can be represented.
Each patch is of 16x16 pixel, so for 20 patch representation, total of 5120 neurons used to represent the classification output.
This when compared to traditional deep neural network classification architectures (which uses one-hot-encoded output), only 5120 number of classes would be represented.

This work extends the PIXEL model to a multi-modal setting for classification task.
As shown in Figure \ref{fig:modeltrain}, the class name as rendered-text is copied along with the input image on an 368x368 image.
The label pixels are masked during training.
The model learns to predict text-pixels matching to the label in the masked pixels/patches.
\textbf{To best of our knowledge this is the first work to use Mask-Image-Modeling (MIM) with multi-modal learning for classification task to create a close-to number-of-classes-agnostic DNN model architecture.}

Summary of contributions of this paper:
\begin{itemize}
  \item Developed a multi-modal training and inference framework \textbf{OneCAD} for classification task using PIXEL model (or any MAE-vision-transformer model).
  \item Built a close-to class-agnostic DNN model architecture for classification task using OneCAD.
  \item Showed that MIM task with multi-modal learning can be used for classification task
  \item Applied the proposed method to both natural and medical domain images.
\end{itemize}

\begin{figure*}[]
\vskip 0.1in
\begin{center}
\centerline{\includegraphics[scale=0.7]{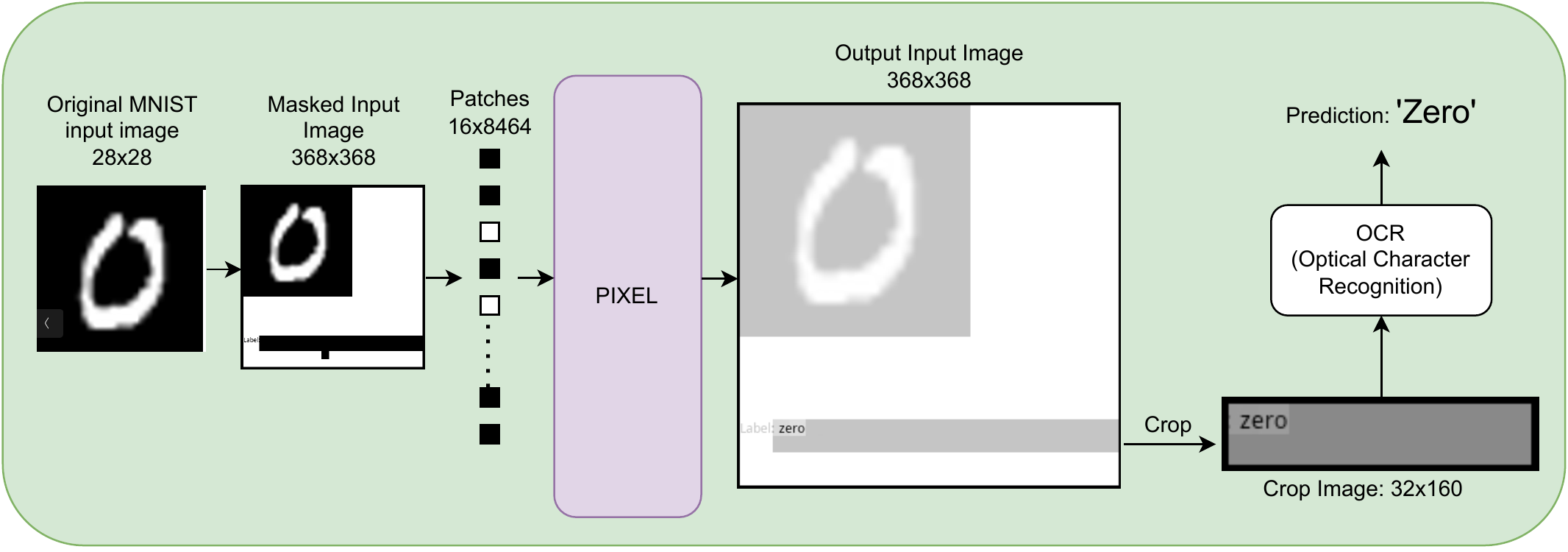}}
\caption{The figure shows our proposed OneCAD's Inference loop. The input is copied on 368x368 image. The label pixels are masked. The PIXEL model takes this masked input image and tries predicts text-pixels matching to the label. OCR is used to convert the visual-text to text label. The text label is compared to the original label for accuracy calculations.}
\label{fig:modelinfer}
\end{center}
\vskip -0.1in
\end{figure*}

\section{Related Work}

\textbf{Image to text non-generative models}:
CLIP \cite{shen2021much}  model can be used as an universal classifier but with several limitations.
By converting class labels to a caption of format "A photo of a \{object\}", contrastive training can be used to perform classification task.
This allows model to learn any number of classes without any model architecture change.
Drawback of such models is that, the generated label embeddings are vocabulary dependent.
Also, the mapping of these output label embeddings to the label text needs to be maintained.
In contrast to the CLIP, in our proposed work, the class is represented by text-pixels, which are directly interpretable.
Moreover, pixel representations of text are vocabulary-free.



\textbf{Image to text generative models}: 
Pix2Struct \cite{lee2022pix2struct} model if finetuned for generating text for classification label, can be used as a universal classifier with similar limitations as CLIP.
In general, any vision-encoder-text-decoder models can also be used as an universal classifier with similar limitations as CLIP.
The vision-encoder (like ViT \cite{dosovitskiy2020image}, Swin \cite{liu2021swin} and DeiT \cite{touvron2021training}) can be used to encode the image into latent embeddings. 
These embeddings can then be passed to a decoder (like GPT2 \cite{radford2019language}) to auto-regressively generate label (class name).

\textbf{Multimodal and Visual text}: The models which use text rendered on image are vocabulary-free models. These model are an attempt towards removing the vocabulary bottleneck.
PIXEL \cite{rust2022language} was inspired from \cite{salesky2021robust}. 
It is trained to perform language modeling from image-pixels. 
The input text is rendered on image-pixels and given as an input to the model.
Model reconstructs the input image to predict masked text to perform multiple language tasks.
CLIPPO \cite{tschannen2022image} is a multi-modal learning model. 
Natural-image and text-image are given as inputs to the model.
The text-image is a rendered image containing text (image-caption) in it. 
These two separate images are provided to the model to perform vision and language tasks.

\textbf{Multimodal}: 
DONUT \cite{kim2021ocr} learns text in the input image in various formats to peform documents understanding tasks. Other such multi-modal models involving text are
Dessurt \cite{davis2023end},
GIT2 Image-to-text generation\cite{wang2022git},
Pali \cite{chen2022pali},
CLIP \cite{shen2021much}, 
ALIGN\cite{jia2021scaling}, 
LIMoE\cite{mustafa2022multimodal}, 
Zero-shot text-to-image generation \cite{ramesh2021zero},
and Pix2Struct \cite{lee2022pix2struct}.

\section{Methodology} 
\label{section:methodology}

\textbf{Problem}: The model architectures used for classification task are dependent on the number of classes in the dataset it is trained on.
Usually, the last linear layer is these architectures are changed to adapt to the new dataset with different number of classes.
This dependence is due to one-hot encoding of the classes.
When these architectures are scaled for learning large number of classes, the number of parameters required in the last linear layer becomes enormous. 
For example, if model's last feature size (CNNs or Transformers) is 4096 and the number of classes in a dataset are 10 Million, then the number parameters required in the last linear layer alone is $\sim$40 billion parameters.
So, to create an universal classifier architecture which can classifiy any number of classes, these architectures come at a high computational and memory cost.
So there is a need for number-of-class-agnostic model architecture for classification task.

\textbf{Solution}: This work proposes a solution to create a (close to) number-of-class-agnostic model architecture for classification task.
The one-hot encoding of the classes is replaced by representation of the label (class name text) in pixels.
Representing label text in form of image-pixels allows a large number of labels to be represented in limited number of pixels/patches.
For example, if the label text occupies total of 10 patches, and each patch is of 16x16, then total of 2560 image-pixels outputs represent one label.
The traditional classification-models with 2560 outputs would only predict 2560 number of classes.
Now, if each output-image-patch can represent one english-letter, then 10 such patches in sequence allows it to represent more than 100 trillion words/label-text.
Increasing number of label text patches allows the model to represent very large number of classes. 
This essentially allows model to be agnostic to the number of classes in the dataset it is trained on, making it a close-to number-of-class-agnostic model architecture.

This work proposes \textbf{OneCAD} (One Classifier for All image Datasets) training and inference framework to achieve this as shows in Figure \ref{fig:modeltrain}.
It uses PIXEL model with multi-modal input and Masked-Image-Modeling training task to achieve creating a number-of-class-agnostic model architecture.
Detailed explanation for model architecture, training procedure and inference steps are given in section \ref{subsec:modelarchitecture}, \ref{subsec:training} and \ref{subsec:inference}.



\subsection{Model Architecture}
\label{subsec:modelarchitecture}

Base PIXEL model \cite{rust2022language} is used.
PIXEL is built using ViT-MAE (Masked Auto Encoding Visual Transformer) \cite{he2022masked}.
This model performs language related tasks using rendered text images.
In the original paper, the model is trained for language tasks.
This work, extends PIXEL to a multi-modal setting. 
Image and rendered text is combined on one image to create an multi-modal input as shown in Fig. \ref{fig:modeltrain}.
For all the datasets used in this work, the images from datasets are resized to 224x224 and then copied on an 368x368 image.
Label text is then written on 368x368 image using PIL python package.
The experiments demonstrate that it possible to train and predict text on images obtained from this simple python packages, helping to avoid expensive rendering used in the original work.
\textit{The entire row of patches containing label-text is masked during training and inference.}
Similar to the original PIXEL work, the 368x368 input image is converted to 16x8464 image and then given to the model for training and inference.

\textbf{When running classification task on different datasets, there is no change in model architecture. One single architecure is used for MNIST (10 classes), CIFAR10 (10 classes), CIFAR100 (100 classes) and COVIDx (2 classes), since our proposed framework is number-of-class-agnostic. To learn new class labels, only retraining (partial or full) is required and no model architecture change needed.}

\subsection{Training}
\label{subsec:training}
Fig. \ref{fig:modeltrain} shows the training framework.
Mask-Image-Modeling (MIM) task is used for training the PIXEL model to predict label text on the output image.
Full image is reconstructed at the output by the model.
The masked pixels which have label-text in them are predicted during training.
The unmasked input-image is used as the desired output.
Mean-squared-error loss is used as loss function during training.

\subsection{Inference/evaluation}
\label{subsec:inference}
Fig. \ref{fig:modelinfer} shows the inference/testing framework.
During inference, the masked input image is given to the model.
The model tries to predict the label text at the masked patches.
These predicted patches are cropped, so a 32x160 image containing the predicted text is provided to the OCR.
OCR (Optical Character Recognition) engine/model is used to for converting the visual-text on image to a text-string.
This paper, uses EasyOCR python package as an OCR model for inference.
The OCR model takes 32x160 image as input and outputs text-string present in the input image.
This text-string is compared with the label (class name) string for accuracy calculations.

\section{Experiments}

\textbf{Experimental setup}: 
PIXEL's base model is used for all the results. 
It has a Vision-Transformer (ViT) and a light weight decoder which generates output image.
The model has 112 million parameters.
The input to the model is 368x368 image which contains Image and text in it.
Patch size of 16x16 is used. So total patches are 529.
Unlike the original PIXEL work which uses rendering to put text on image, this paper demonstrates that it is possible to learn with simple PIL python generated images for text.
The training uses AdamW as optimizer, cosine learning rate scheduler with warm-up steps (5\% of maximum iterations), maximum learning rate is 5e-6 and minimum learning rate is set to 5e-7.
Batch size of 16 is used for all datasets.
One 3090 GPU was used for all the training and testing.
Results are shown on both and natural images (MNIST, CIFAR10, CIFAR100) and medical image data (COVIDx).
Accuracy is used as evaluation metric.
FW (Full Word) accuracy represents the accuracy when all the predicted characters match all the label characters.
FC (First character) accuracy represents the accuracy when the first character of the prediction and label are compared. 
FTC (First two characters) accuracy represents the accuracy when the first two characters of the prediction and label are compared.
FC and FTC accuracy metrics are introduced due to limitations of OCR. Detailed explanation of this limitation provided in section \ref{subsec: acc-limit}.
When running classification task on different datasets, there is no change in model architecture. 
One single architecure is used for MNIST (10 classes), CIFAR10 (10 classes) and CIFAR100 (100 classes), since our proposed framework is number-of-class-agnostic. 
To learn new class labels, only retraining (partial or full) is required.

\subsection{MNIST}
MNIST is an image dataset of handwritten digits from 0 to 9. 
It contains 60K training images and 10K test images.
The image sizes are 28x28.
The image is resized to 224x224 and along with the label are embedded onto 368x368 image as shown in Fig. \ref{fig:modeltrain}.

After 10 epochs of training, the trained model achieves only 69\% accuracy.
This is partially due to different brightness of the predicted text.
It was observed that, if the label is correctly predicted but its pixel-brightness varies, then the OCR is not able to detect text correctly.
This issue is highlighted in section \ref{subsec: acc-limit}.
Brightness of the output text is important for OCR to work with high accuracy.
Changing brightness and making it darker by factor of 0.7 leads to higher accuracy of 87.66\%.
Calculating accuracy over first two predicted characters further improves the accuracy by ~0.3\%.
Table \ref{table:mnist-acc} shows both FW and FTC accuracies.
More training will help improve the accuracy further. 
Also if the PIXEL model is pretrained for multi-modal input, it will lead to further improvement in the accuracy.
Work is in progress to train the model for longer epochs, obtain multi-modal-pretrained pixel model and a better OCR model.

\begin{table}[h]
\caption{Classification accuracies (on testset) of PIXEL model when trained on full MNIST dataset}
\label{table:mnist-acc}
\vskip 0.1in
\begin{center}
\begin{small}
\begin{sc}
\begin{tabular}{lcccr}
\toprule
FW & FTC & Brightness & Accuracy(\%) \\
\midrule
$\surd$   &  $\times$   & 1.0  & 69.00$\pm$ 0.5 \\
$\surd$   &  $\times$   & 0.7  & 87.66$\pm$ 0.25 \\
$\times$   &  $\surd$   & 0.7  & 87.91$\pm$ 0.25\\

\bottomrule
\end{tabular}
\end{sc}
\end{small}
\end{center}
\vskip -0.1in
\end{table}

\subsection{CIFAR10}

CIFAR10 dataset contains color images of size 32x32. 
Total of 50K training and 10K testing images are provided.
Number of classes are 10. 
These images are resized to 224x224 and along with text label is copied on 368x368 image as shown in Figure \ref{fig:modeltrain}.
The model architecture used for CIFAR10, is exactly same as used for MNIST dataset.

\begin{figure}[h]
\vskip 0.1in
\begin{center}
\centerline{\includegraphics[width=\columnwidth]{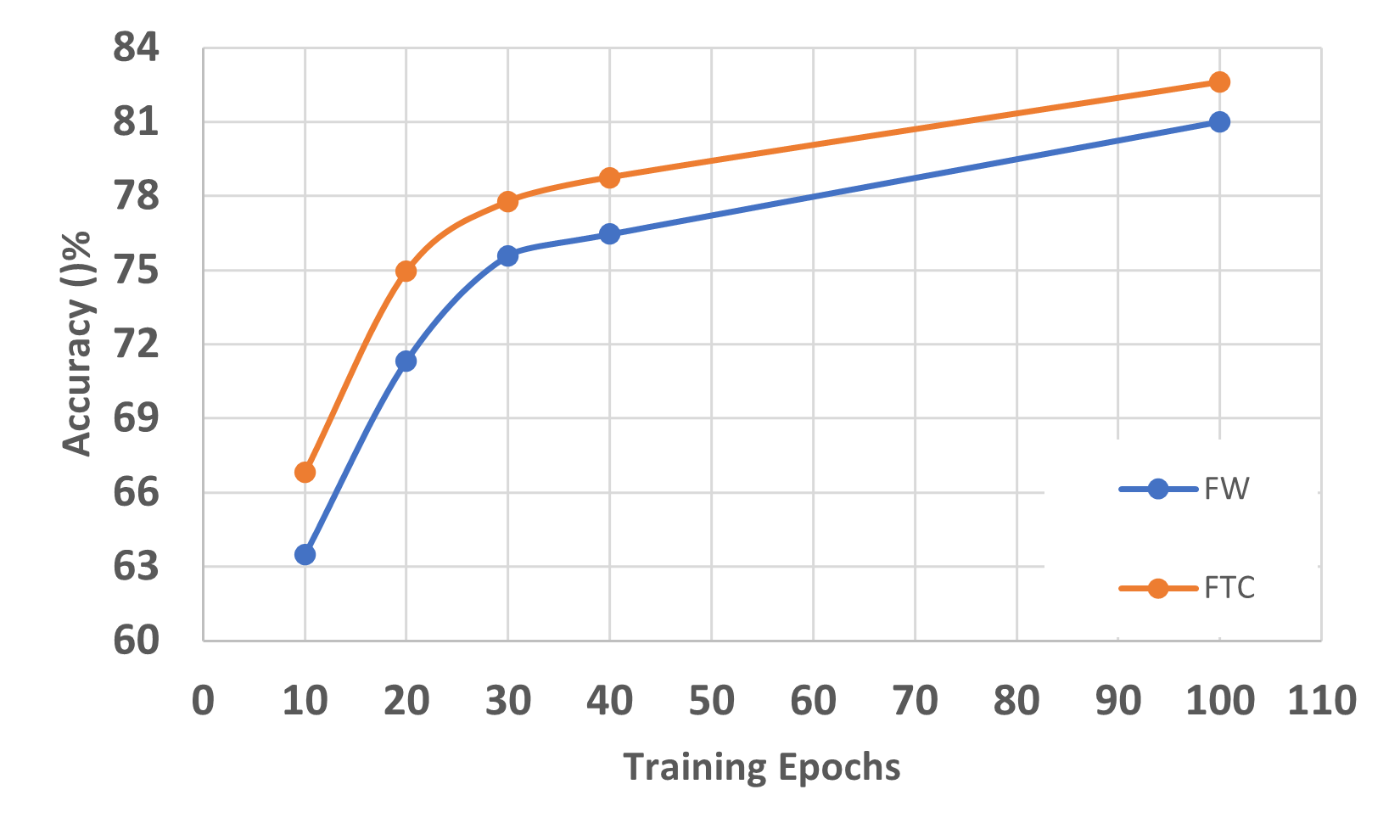}}
\caption{Epoch vs Accuracy graph for CIFAR10 dataset. Accuracies are calculated on test set containing 10K images. FW accuracy is when full predicted word is compared with label word. FTC accuracy is calculated by comparing first two characters of predicted text with the label's first two characters.}
\label{fig:cifar10-e-vs-acc}
\end{center}
\vskip -0.1in
\end{figure}

The model is trained on 100 epochs and the test accuracy curves are shown in Figure \ref{fig:cifar10-e-vs-acc}.
As the training epochs increase the test accuracy improves.
81.01\% accuracy (FW accuracy) is achieved by model when predicted full text word is compared to the full text label.
82.6\% is achieved when first two characters of predicted and label texts are compared as seen in Table \ref{table:cifar10-acc}.
It can be trained further and has potential to achieve state of the art results.
Surprisingly, if the input image is normalized, the accuracy values degrade by over 4\%.
The likely reason is that the pretrained PIXEL model was not trained with normalized input images.
The original PIXEL is pretrained on gray scale rendered text images.
Pretraining the PIXEL model with multi-modal input similar to ours will help to boost the accuracy further.
Lowering the brightness from 1.0 to 0.7 of the 32x160 cropped output image of predicted text, helps OCR predict better as observed on other datasets as well.

\begin{table}[h]
\caption{Classification accuracy on testset of CIFAR10 dataset}
\label{table:cifar10-acc}
\vskip 0.15in
\begin{center}
\begin{small}
\begin{sc}
\begin{tabular}{lcccr}
\toprule

FW & FTC & Brightness & Accuracy(\%) \\
\midrule
$\surd$   &  $\times$   & 0.7  & 81.01 $\pm$0.25 \\
$\times$   &  $\surd$   & 0.7  & 82.60 $\pm$0.25\\

\bottomrule
\end{tabular}
\end{sc}
\end{small}
\end{center}
\vskip -0.1in
\end{table}


\subsection{CIFAR100}

CIFAR100 dataset contains 'RGB' (color) images of size 32x32. 
Total of 50K training and 10K testing images are provided.
Number of classes are 100. 
These images are resized to 224x224 and along with text label is copied on 368x368 image as shown in Figure \ref{fig:modeltrain}.
The model architecture used for CIFAR100 (100 classes), is exactly same as used for CIFAR10 (10 classes) and MNIST dataset (10 classes).

\begin{figure}[h]
\vskip 0.1in
\begin{center}
\centerline{\includegraphics[width=\columnwidth]{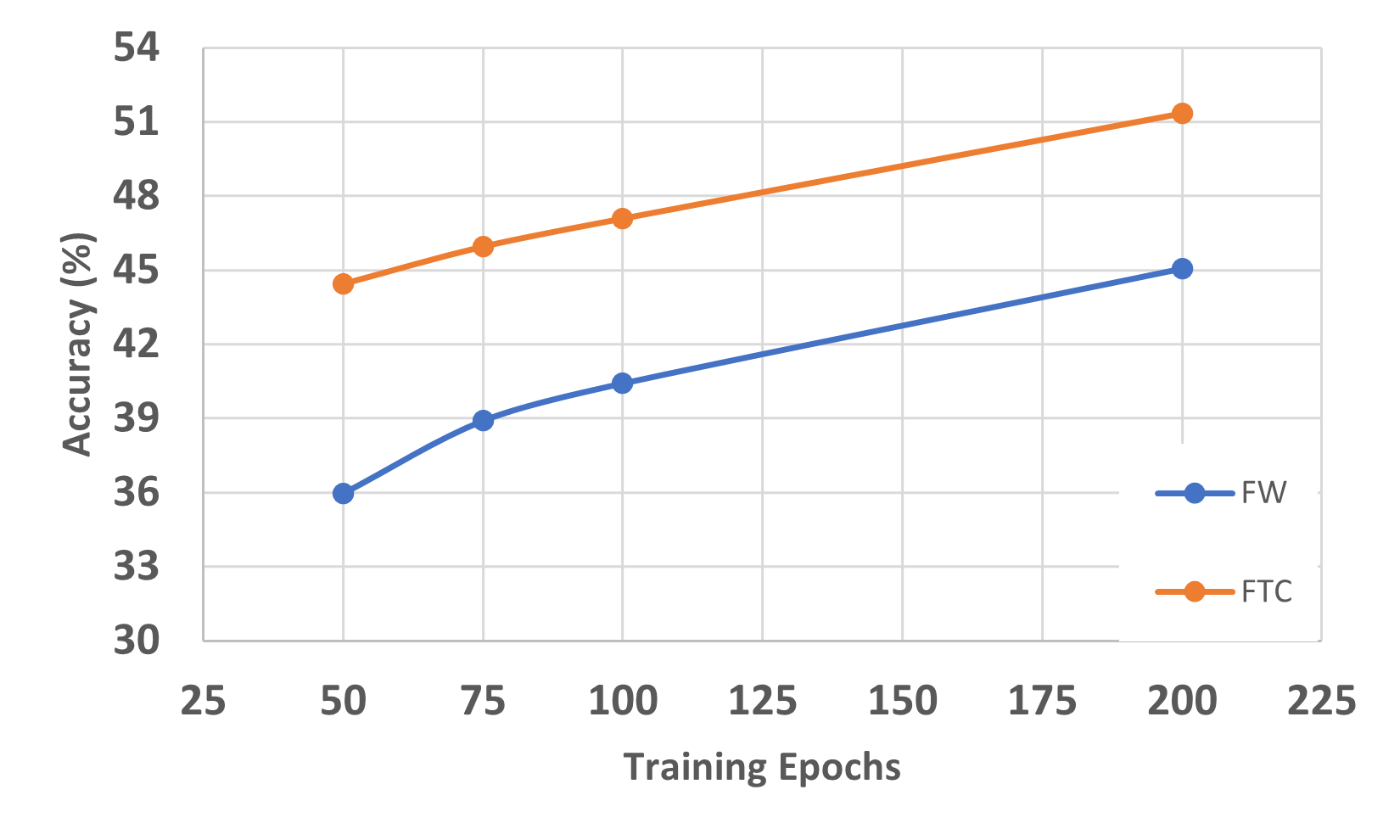}}
\caption{Epoch vs Accuracy graph for CIFAR100 dataset. Accuracies are calculated on test set containing 10K images. FW accuracy is when full predicted word is compared with label word. FTC accuracy is calculated by comparing first two characters of predicted text with the label's first two characters.}
\label{fig:cifar100-e-vs-acc}
\end{center}
\vskip -0.1in
\end{figure}

More training epochs are required to train CIFAR100 dataset as compared to CIFAR10. 
With 100 training epochs, the model achieves 40.42 \% accuracy (FW accuracy) and with 200 epochs it achieves 45.06\%.
It takes $\sim$4 days to train on CIFAR100 dataset for 200 epochs using one 3090 GPU.
Long training time is due to the large input size of 368x368.
Accuracy of the model can further be improved by training it for more number of epochs and pretraining the PIXEL model with multi-modal input.
FW and FTC accuracies are shown in Table \ref{table:cifar100-acc}.

\begin{table}[h]
\caption{Classification accuracy on testset of CIFAR100 dataset}
\label{table:cifar100-acc}
\vskip 0.15in
\begin{center}
\begin{small}
\begin{sc}
\begin{tabular}{lcccr}
\toprule

FW & FTC & Brightness & Accuracy(\%) \\
\midrule
$\surd$   &  $\times$   & 0.7  & 45.06 $\pm$0.25 \\
$\times$   &  $\surd$   & 0.7  & 51.34 $\pm$0.25\\

\bottomrule
\end{tabular}
\end{sc}
\end{small}
\end{center}
\vskip -0.1in
\end{table}

\subsection{COVIDx}
This dataset consists of chest x-ray images for detection of COVID-19 \cite{wang2020covid}.
It contains over 30K images. 
This paper uses latest version 7 obtained from kaggle. 
The task is binary classification task i.e, classifiying image into COVID-19 'positive' or 'negative'.
Similar to  \cite{NEURIPS2022_d925bda4}, the train dataset is split to 90\% for training set and 10\% for validation set.
Testing is done using test images provided in the dataset.
Accuracy is used as evaluation metric.
The FC accuracy metric is introduced because the predictions for all letters are not always clear.
Also the OCR model during inference does not always recognize the words due to text brightness issues.
The model architecture used for COVIDx (2 classes), is exactly same as used for CIFAR100 (100 classes), CIFAR10 (10 classes) and MNIST dataset (10 classes).

The PIXEL network is not trained on any image dataset except the rendered text images.
The network is able to achieve good performance without pretraining it on large medical imaging dataset like MIMIC-CXR dataset \cite{johnson2019mimiccxrjpg}.
The PIXEL network is trained for total 40 epochs with varying train dataset sizes and the results are shown in Table \ref{table:covid-data-per}. 


\begin{table}[h]
\caption{Classification accuracy vs Amount of training data in percentage of COVIDx-v7 dataset}
\label{table:covid-data-per}
\vskip 0.15in
\begin{center}
\begin{small}
\begin{sc}
\begin{tabular}{lcccr}
\toprule
Data(\%) & FW(\%) & FC(\%) \\
\midrule
1   &  0.50 $\pm$0.5    & 51.74 $\pm$0.5 \\
10  &  78.50 $\pm$0.5   & 83.75 $\pm$0.5 \\
100 &  90.50 $\pm$0.25   & 92.00 $\pm$0.25 \\

\bottomrule
\end{tabular}
\end{sc}
\end{small}
\end{center}
\vskip -0.1in
\end{table}


\section{Observations and interpretations}

\subsection{Accuracy limitations due to OCR and model}
\label{subsec: acc-limit}
\begin{figure}[h]
\vskip 0.1in
\begin{center}
\centerline{\includegraphics[width=\columnwidth]{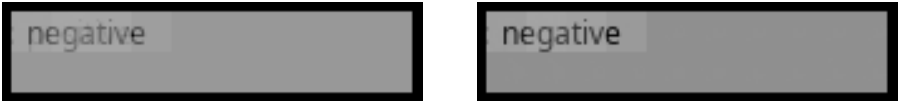}}
\caption{Output samples}
\label{fig:brightness}
\end{center}
\vskip -0.1in
\end{figure}

Brightness of the predicted text greatly effects the ability of OCR to recognize characters.
The predicted text labels for COVIDx dataset are 'positive' and 'negative'.
The brightness of these predictions are not always consistent.
As seen in Fig. \ref{fig:brightness}, human eye can recognize the text in the left box as 'negative' label, but the OCR fails to recognize the characters and leads to degradation of performance.

Two ways used in this work to overcome this problem are, first is to change brightness of the predicted text image and, second is to only consider first few characters when calculating accuracy.
Making pixels slightly darker helps the OCR to predict better.
Reducing brightness from 1.0 to 0.7 greatly improves the accuracy is seen in the table \ref{table:bright value} for the MNIST datastet.
Considering only the first letter of the predictions i.e, 'p' or 'n' for COVIDx dataset largely improves the accuracy of the model. 
Accuracy gain of 2\% for COVIDx dataset can observed as seen in Table \ref{table:bright value}.

\begin{table}[h]
\caption{Classification accuracies of PIXEL model when trained on full various dataset}
\label{table:bright value}
\vskip 0.1in
\begin{center}
\begin{small}
\begin{sc}
\begin{tabular}{lcccr}
\toprule
Data & FW & FTC/FC & Bright. & Acc (\%) \\
\midrule
MNIST & $\surd$   &  $\times$   & 1.0  & 69.00 \\
MNIST & $\surd$   &  $\times$   & 0.7  & 87.66 \\
MNIST & $\times$   &  \textbf{$\surd$}   & \textbf{0.7}  & \textbf{87.91} \\
COVIDx & $\surd$   &  $\times$  & 0.7  & 90.50 \\
COVIDx & $\times$   &  \textbf{$\surd$}  & \textbf{0.7}  & \textbf{92.00} \\

\bottomrule
\end{tabular}
\end{sc}
\end{small}
\end{center}
\vskip -0.1in
\end{table}

Additionally, because the PIXEL model is not pretrained on natural or medical images, the performance of the model is not state-of-the-art. 
Pretraining and fine-tuning PIXEL model with multi-modal inputs will help it achieve competitive results on classification task.

\section{Conclusion and Future direction}
This work demonstrates how a close-to number-of-class-agnostic model architecture can be built using our proposed OneCAD framework.
PIXEL model with multi-modal input and Mask-Image-Modeling task is used for classification task.
This work demonstrates the proof-of-concept of creating a number-of-class-agnostic model for classification task with preliminary results on MNIST, CIFAR10, CIFAR100 and COVIDx dataset.
The performance can be improved by further training, using a multi-modal-pretrained PIXEL model and using a better OCR model.
Work is in progress to obtain multi-modal-pretrained PIXEL model and training models for more epochs to achieve state of the art results on classification task.


\nocite{langley00}

\bibliography{biblography}
\bibliographystyle{icml2023}




\end{document}